\pdfoutput=1

\documentclass[11pt]{article}

\usepackage[]{acl}
\usepackage{amsmath}
\usepackage{comment}
\usepackage{booktabs}
\usepackage{pifont}
\usepackage{times}
\usepackage{latexsym}
\usepackage[edges]{forest}
\usepackage[T1]{fontenc}

\usepackage[utf8]{inputenc}

\usepackage{microtype}

\usepackage{inconsolata}

\definecolor{myblue}{RGB}{112, 156, 156}
\definecolor{lightblue}{RGB}{237, 250, 252}

\title{Beyond Chain-of-Thought: A Survey of Chain-of-X Paradigms for LLMs}

\author{Yu Xia$^{1,2}$ \quad Rui Wang$^{3}$ \quad Xu Liu$^{1}$ \quad Mingyan Li$^{1}$ \quad Tong Yu$^{4}$ \\ {\bf Xiang Chen$^{4}$ \quad Julian McAuley$^{2}$ \quad Shuai Li$^{1}$\thanks{\; Corresponding author.}}\\
$^1$Shanghai Jiao Tong University \quad 
$^2$UC San Diego  \quad
$^3$Duke University \quad
$^4$Adobe Research \\
\texttt{\{yux078, jmcauley\}@ucsd.edu} \qquad
\texttt{\{tyu, xiangche\}@adobe.com}  \\
\texttt{rui.wang16@duke.edu} \quad
\texttt{\{liu\_skywalker, QYLJM1217, shuaili8\}@sjtu.edu.cn} 
}

\begin{document}

\maketitle

\begin{abstract}
Chain-of-Thought (CoT) has been a widely adopted prompting method, eliciting impressive reasoning abilities of Large Language Models (LLMs). 
Inspired by the sequential thought structure of CoT, a 
number of Chain-of-X (CoX) methods have been developed to address challenges across diverse domains and tasks. 
In this paper, we provide a comprehensive survey of Chain-of-X methods for LLMs in different contexts.
Specifically, we categorize them by taxonomies of nodes, i.e., the X in CoX, and application tasks. 
We also discuss the findings and implications of existing CoX methods, as well as potential future directions. 
Our survey aims to serve as a detailed and up-to-date resource for researchers seeking to apply the idea of CoT to broader scenarios.
    
\end{abstract}

\section{Introduction}

Large Language Models (LLM) have shown strong reasoning capabilities when prompted with the Chain-of-Thought (CoT) method \cite{wei2022chain}. 
The essence of CoT is to decompose complex problems into sequences of intermediate subtasks \cite{chu2023survey}.
By handling these subtasks step by step, LLMs are able to focus on important details and assumptions, which substantially improves the performance across a wide range of reasoning tasks \cite{huang-chang-2023-towards, chu2023survey}. 
CoT's intermediate steps can also offer a more transparent reasoning process, facilitating easier interpretation and evaluation of LLMs \cite{yu2023towards}.

With the success of CoT, a number of Chain-of-X (CoX) methods have subsequently been developed \cite{yu2023chain}. 
Extending beyond reasoning thoughts, recent CoX methods have constructed the chain with various components, such as Chain-of-Feedback \cite{lei2023chain, dhuliawala2023chain}, Chain-of-Instructions \cite{zhang2023coie, anugrah2024chain}, Chain-of-Histories \cite{luo2024chain, xia2024enhancing}, etc. 
These methods have been applied to diverse tasks involving LLMs beyond reasoning, including multi-modal interaction \cite{xi2023chain, zhang2024speechgpt}, hallucination reduction \cite{lei2023chain, dhuliawala2023chain}, planning with LLM-based agents \cite{zhan2023you, zhang2024android}, etc.

\paragraph{Motivation}
Despite their growing prevalence, these CoX methods have not yet been collectively examined or categorized, leaving a gap in our understanding of their potential. This survey offers a structured overview capturing CoX's essence and diversity for further exploration and innovation.

\begin{figure*}[!t]
    \centering
    \includegraphics[width=0.84\linewidth]{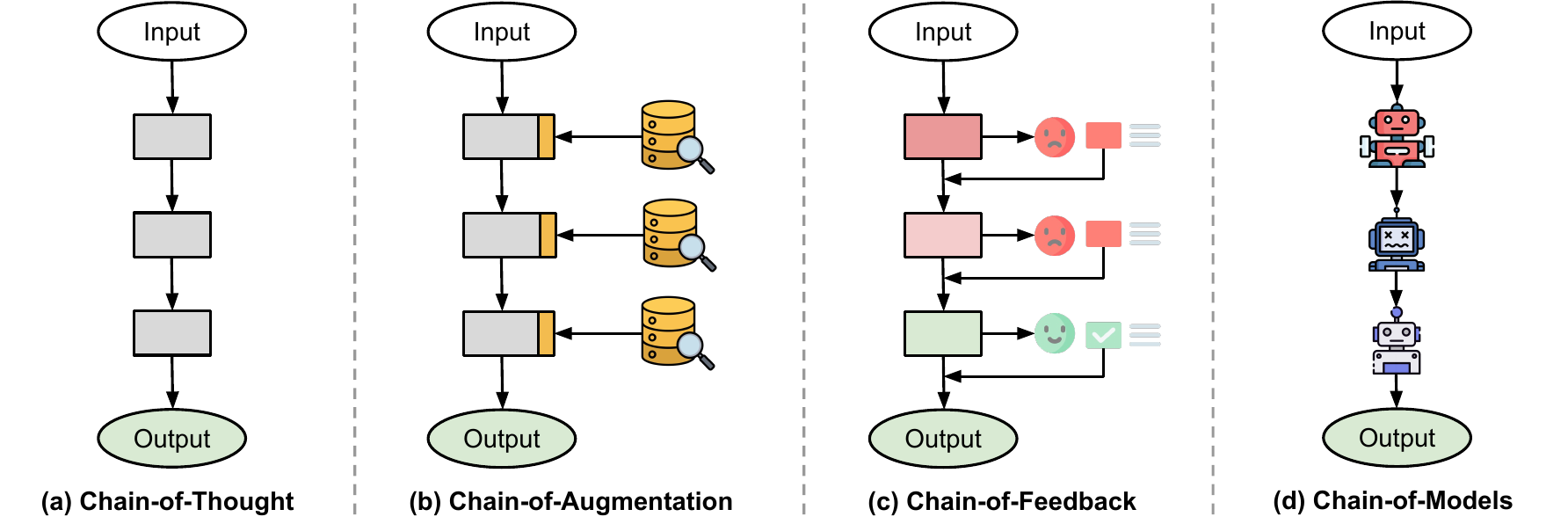}
    \caption{Illustrations of Chain-of-X paradigms with four types of nodes: (a) Intermediates, e.g., Thought (\S \ref{sec:inter}), (b) Augmentation (\S \ref{sec:augmt}), (c) Feedback (\S \ref{sec:fdb}), and (d) Models (\S \ref{sec:models}).}
    \vspace{-0.5em}
    \label{fig:cox}
\end{figure*}

\paragraph{Distinguishing Focus}
While several surveys have explored CoT \cite{chu2023survey, yu2023towards, besta2024topologies}, they focus primarily on the reasoning thoughts of different structures, e.g., Chain-of-Thought as illustrated in Figure \ref{fig:cox}(a).
In contrast, this paper focuses on the multifaceted component designs of Chain-of-X beyond reasoning thoughts as shown in Figure \ref{fig:cox}, offering insights of the CoT concept in broader domains.
We present a comprehensive review by taxonomies of the X in CoX and tasks to which these methods are applied.

\paragraph{Overview of the Survey} 
We first provide background information on Chain-of-Thought and define Chain-of-X as its generalization (\S \ref{sec:preliminary}).
Next, we categorize CoX methods by the types of components used to construct the chains (\S \ref{sec:nodes}).
Furthermore, based on the application areas of these CoX methods, we categorize them by tasks (\S \ref{sec:tasks}).
Then, we discuss insights from existing CoX methods and explore potential future directions (\S \ref{sec:discussion}).
A detailed structure of the survey is presented in Figure \ref{fig:survey}.

\section{What is Chain-of-X?}\label{sec:preliminary}

In this section, we introduce some background information about Chain-of-Thought prompting and then define a generalized concept of Chain-of-X.

\paragraph{Chain-of-Thought}

CoT prompting is a methodology that substantially enhances the reasoning capabilities of LLMs. 
Introduced by \citet{wei2022chain}, CoT involves prompting LLMs with a structured format of \texttt{<input, thoughts, output>}, where `thoughts' encompass coherent and intermediate natural language reasoning steps leading to the final answer. 
CoT's effectiveness is most pronounced in tasks that require complex reasoning. 
Traditional few-shot learning methods often falter in such scenarios, as they tend to provide direct answers without rationales. 
In contrast, CoT prompting excels by breaking down a complex task into manageable intermediate steps.
These steps guide the model through a logical progression, enhancing its capability to tackle complex problems such as arithmetic, commonsense, and symbolic reasoning \cite{wang2023selfconsistency, lyu-etal-2023-faithful}.
Additionally, \citet{kojima2022large} have also demonstrated strong performance of zero-shot CoT by prompting ``\textit{Let's think step by step.}''.
The explicit reasoning steps also provide a transparent pathway for the model's thought process, allowing for further evaluations and corrections \cite{yu2023towards}.

\begin{figure*}[t!]
\centering
\begin{forest}
for tree={   
font=\fontsize{6}{5}\selectfont,
draw=myblue, semithick, rounded corners,
       minimum height = 1.ex,
        minimum width = 2em,
    anchor = west,
     grow = east,
forked edge,        
    s sep = 0.5mm,    
    l sep = 2mm,    
 fork sep = 1mm,    
           }
[A Survey of Chain-of-X, rotate=90, anchor=center
    [Taxonomy of Tasks (\S \ref{sec:tasks}), fit=band, text width=0.9cm
        [Evaluation Tools (\S \ref{sec:eval}), text width=2.61cm, l sep = 2mm
            [{E.g., CoUtterances \cite{bhardwaj2023red}, BadChain \cite{xiang2024badchain}, CoFeedback \cite{ahn2024chain}}, text width=10.6cm, fill=lightblue]
        ]
        [LLMs as Agents (\S \ref{sec:agent}), text width=2.61cm, l sep = 2mm
            [{E.g., CoAction$^a$ \cite{zhan2023you}, CoActionThought \cite{zhang2024android}, CoContacts \cite{xiao2024unified}}, text width=10.6cm, fill=lightblue]
        ]
        [Instruction Following (\S \ref{sec:instr_fl}), text width=2.61cm, l sep = 2mm
            [{E.g., CoTask \cite{li2023ecomgpt}, LogiCoT \cite{liu-etal-2023-logicot}, CoImagination \cite{zhou2024minedreamer}}, text width=10.6cm, fill=lightblue]
        ]
        [Multi-Step Reasoning (\S \ref{sec:ms_reas}), text width=2.61cm, l sep = 2mm
            [{E.g., CoDensity \cite{adams-etal-2023-sparse}, CoLogic \cite{servantez2024chain}, CoEvent \cite{bao2024chain}}, text width=10.6cm, fill=lightblue]
        ]
        [Factuality \& Safety (\S \ref{sec:fact}), text width=2.61cm
            [Alignment, text width=1.35cm, l sep = 1mm
                [{E.g., CoUtterances \cite{bhardwaj2023red}, CoHindsight \cite{liu2024chain}}, text width=8.9cm, fill=lightblue]
            ]
            [Hallucination, text width=1.35cm, l sep = 1mm
                [{E.g., CoVerification \cite{dhuliawala2023chain}, CoKnowledge$^a$ \cite{li2023chain_k}}, text width=8.9cm, fill=lightblue]
            ]
        ]
        [Multimodal Interaction (\S \ref{sec:mm_inter}), text width=2.61cm
            [Text-Speech, text width=1.35cm, l sep = 1mm
                [{E.g., CoInformation \cite{zhang2024speechgpt}, CoModality \cite{zhang-etal-2023-speechgpt}}, text width=8.9cm, fill=lightblue]
            ]
            [Text-Code, text width=1.35cm, l sep = 1mm
                [{E.g., CoRepair \cite{wang2023intervenor}, CoCode \cite{li2023chain}, CoSimulation \cite{la2024code}}, text width=8.9cm, fill=lightblue]
            ]
            [Text-Table, text width=1.35cm, l sep = 1mm
                [{E.g., CoCommand \cite{zha2023tablegpt}, CoTable \cite{wang2023chain}}, text width=8.9cm, fill=lightblue]
            ]
            [Text-Image, text width=1.35cm, l sep = 1mm
                [{E.g., MMCoT \cite{zhang2023multimodal}, CoLook \cite{xi2023chain}, CoSpot \cite{liu2024chain-a}}, text width=8.9cm, fill=lightblue]
            ]
        ]
    ]
    [Taxonomy of Nodes (\S \ref{sec:nodes}), fit=band, text width=0.9cm
        [Chain-of-Models (\S \ref{sec:models}), text width=2.61cm, l sep = 2mm
            [{E.g., CoExperts \cite{xiao2024chainofexperts}, ChatEval \cite{chan2024chateval}, CoDiscussion \cite{tao2024chain}}, text width=10.6cm, fill=lightblue]
        ]
        [Chain-of-Feedback (\S \ref{sec:fdb}), text width=2.61cm
            [Self Feedback, text width=1.35cm, l sep = 1mm
                [{E.g., Self-Refine \cite{madaan2023selfrefine}, CoVerification \cite{dhuliawala2023chain}}, text width=8.9cm, fill=lightblue]
            ]
            [Ext. Feedback, text width=1.35cm, l sep = 1mm
                [{E.g., Chain-of-3DThought \cite{yamada2024l3go}, CoHindsight \cite{liu2024chain}}, text width=8.9cm, fill=lightblue]
            ]
        ]
        [Chain-of-Augmentation (\S \ref{sec:augmt}), text width=2.61cm
            [Others, text width=1.35cm, l sep = 1mm
                [{E.g., CoReference \cite{kuppa2023chain}, CoDictionary \cite{lu2023chain}, CoMemory \cite{hu2023chatdb}}, text width=8.9cm, fill=lightblue]
            ]
            [Tools, text width=1.35cm, l sep = 1mm
                [{E.g., ChatCoT \cite{chen-etal-2023-chatcot}, MultiToolCoT \cite{inaba-etal-2023-multitool}, CoAbstract \cite{gao2024efficient}}, text width=8.9cm, fill=lightblue]
            ]
            [Retrievals, text width=1.35cm, l sep = 1mm
                [{E.g., ReAct \cite{yao2023react}, CoQuery \cite{xu2023search}, CoKnowledge$^a$ \cite{li2023chain_k}}, text width=8.9cm, fill=lightblue]
            ]
            [Histories, text width=1.35cm, l sep = 1mm
                [{E.g., CoOpinion \cite{do2023choire}, CoHistory$^a$ \cite{luo2024chain}, CoHistory$^b$ \cite{xia2024enhancing}}, text width=8.9cm, fill=lightblue]
            ]
            [Instructions, text width=1.35cm, l sep = 1mm
                [{E.g., CoInstructEditing \cite{zhang2023coie}, CoInstructions \cite{anugrah2024chain}}, text width=8.9cm, fill=lightblue]
            ]
        ]
        [Chain-of-Intermediates (\S \ref{sec:inter}), text width=2.61cm
            [Know. Comp., text width=1.35cm, l sep = 1mm
                [{E.g., CoKnowledge$^b$ \cite{wang2023boosting}, CueCoT \cite{wang-etal-2023-cue}, CoSpot \cite{liu2024chain-a}}, text width=8.9cm, fill=lightblue]
            ]
            [Prob. Decomp., text width=1.35cm, l sep = 1mm
                [{E.g., CoT \cite{wei2022chain}, Least-to-Most \cite{zhou2023leasttomost}, CoCode \cite{li2023chain}}, text width=8.9cm, fill=lightblue]
            ]
        ]
    ]
]
\end{forest}
\caption{A survey of Chain-of-X by taxonomies of nodes and tasks (only representative methods are listed due to space limitation and a more complete version can be found in Appendix \ref{sec:app}).}
\vspace{-0.5em}
\label{fig:survey}
\end{figure*}
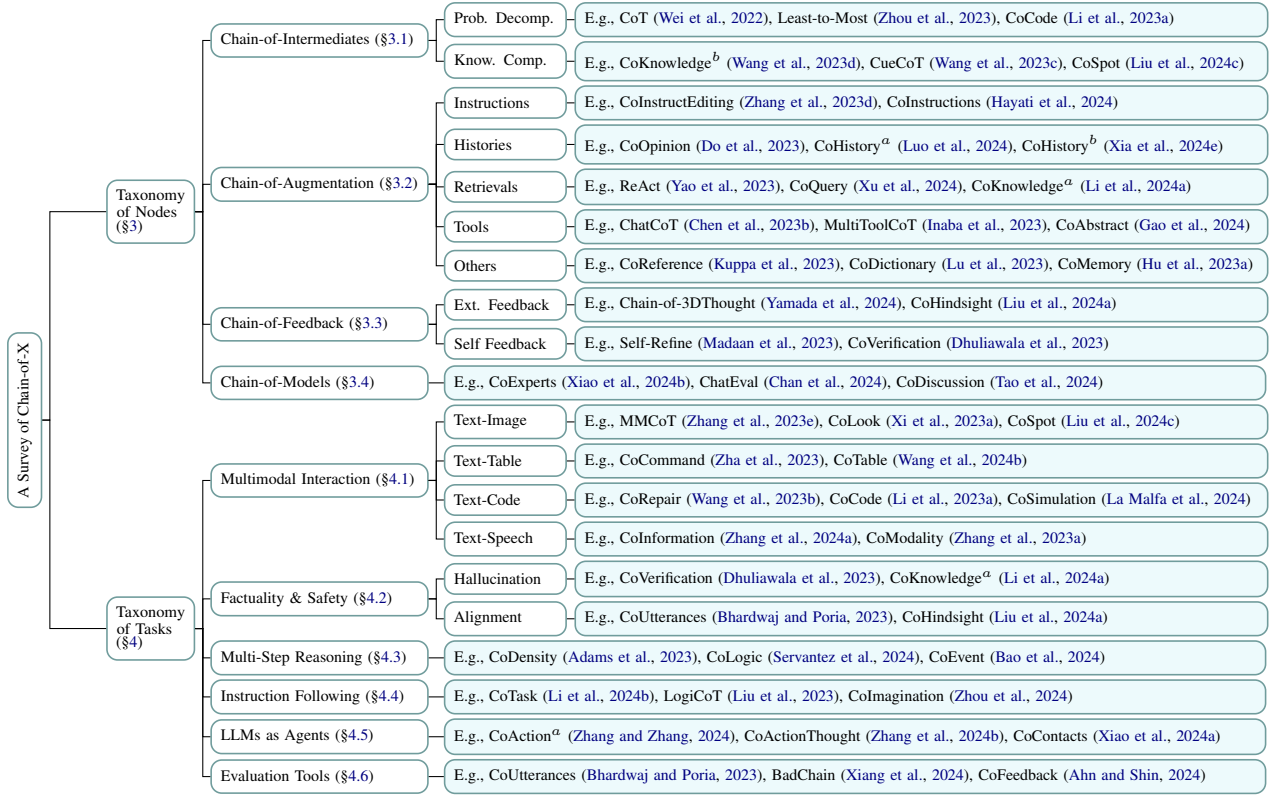

\paragraph{Chain-of-X}
Inspired by the nature of the sequential breakdown, a substantial number of CoX methods have been developed recently \cite{yu2023chain}. 
Here, we define CoX as a generalization of the CoT method for diverse tasks beyond LLM reasoning. 
We refer to the X in CoX as the `node' of the chain structure. 
Beyond the thoughts in CoT prompts, the X in CoX can take various forms tailored to specific tasks, including intermediates (\S \ref{sec:inter}), augmentation (\S \ref{sec:augmt}), feedback (\S \ref{sec:fdb}), and even models (\S \ref{sec:models}), as illustrated in Figure \ref{fig:cox}.
We summarize the types of nodes in existing CoX methods in Figure \ref{fig:survey}.
The idea of CoX is to construct a sequence of problem-related components that either compositionally contribute to the solution or iteratively refine the outputs.
We define a similar structured format for CoX as \texttt{<input, X$_1$, $\dots$, X$_n$, output>} where $n$ is the length of the chain.
Note that this format extends beyond prompting-based strategies like CoT and can be adapted to a variety of algorithmic frameworks or structures for diverse tasks involving LLMs.
For instance, Chain-of-Verification \cite{dhuliawala2023chain} is a hallucination reduction framework that employs an LLM to generate initial responses, composes a sequence of verification questions, and revises its previous responses based on these questions.
In addition to hallucination reduction, CoX methods have been applied to a variety of tasks, as shown in Figure \ref{fig:survey}, including multi-modal interaction (\S \ref{sec:mm_inter}), factuality \& safety (\S \ref{sec:fact}), multi-step reasoning (\S \ref{sec:ms_reas}), instruction following (\S \ref{sec:instr_fl}), LLMs as Agents (\S \ref{sec:agent}), and evaluation tools (\S \ref{sec:eval}).

\section{Chain-of-X Nodes}\label{sec:nodes}
In this section, we survey existing CoX methods by taxonomy of nodes, categorizing them as shown in Figure \ref{fig:survey} based on the distinct nature of the nodes.

\subsection{Chain-of-Intermediates}\label{sec:inter}
Building on the concept of utilizing intermediate steps, a natural evolution of CoT involves generalizing reasoning thoughts to other types of intermediate components. 
Based on the primary focuses, we further divide them into the following subtypes.

\paragraph{Problem Decomposition}
In problem decomposition, the intermediate steps consist of manageable subtasks derived from an original complex problem, which is exemplified by the classic Chain-of-Thought prompting \cite{wei2022chain} for reasoning tasks.
To overcome the challenge of easy-to-hard generalization, \citet{zhou2023leasttomost} further introduce Least-to-Most prompting which breaks down a complex problem into simpler subtasks and solves them in sequence.
Extending beyond natural languages, Chain-of-Code \cite{li2023chain} takes advantage of the syntactic structure and precise computation of code by segmenting a complex task into programmatic subtasks, enhancing the reasoning process through simulated code outputs.
Similarly motivated by computation accuracy, Chain-of-Logic \cite{servantez2024chain} applies a logical decomposition transforming rule-based reasoning tasks into a series of simple logical expressions. 

While such decomposition is widely applied in reasoning tasks, the concept is also echoed in other tasks.
For example, Chain-of-Event \cite{han2024chain} simplifies multi-document summarization into discrete event extraction tasks, significantly enhancing the summarization quality and factuality.
Chain-of-Table \cite{wang2023chain} restructures complex tables into question-specific formats via a sequence of strategic operations, making the data more accessible and tailored to the inquiry.

\paragraph{Knowledge Composition}
In knowledge composition, the primary goal of the intermediate steps is not simplification but the accumulation of relevant information.
This approach aims to enrich the solution with a depth of understanding and details.
For instance, to handle unfactual rationales generated with CoT prompting, \citet{wang2023boosting} propose Chain-of-Knowledge$^b$ to elicit LLMs to generate explicit knowledge evidence at each reasoning step for more grounded question-answering.
Similarly in dialogue systems, CueCoT \cite{wang-etal-2023-cue} collects linguistic cues with intermediate steps to capture contextual user status for more personalized and engaging conversation.

Besides natural language tasks, this technique is also useful in knowledge-intensive visual tasks that require capturing specific visual details.
For example, Chain-of-Spot \cite{liu2024chain-a} and Chain-of-Reasoning \cite{uehara2024advancing} enable vision-language models to focus on key regions of interest, improving reasoning performance with detailed visual evidences.
Likewise, CCoT \cite{mitra2023compositional} utilizes scene-graph representations to extract compositional knowledge from a large multimodal model, which is further used to facilitate its own response generation on vision-language tasks.


\subsection{Chain-of-Augmentation}\label{sec:augmt}
While Chain-of-Intermediates method has proven effective, it falls short when LLMs have limited knowledge for specific tasks or domains. As a result, Chain-of-Augmentation has become a popular variant of CoX, where the chain is augmented with additional knowledge. Based on the types of augmented data, we categorize them as follows.

\paragraph{Instructions}
Given a complex task, determining the next step can be nontrivial for LLMs even with few-shot CoT exemplars, due to misinterpretation or ambiguity \cite{zha2023tablegpt}.
Instructions then serve as an important augmentation, guiding LLMs through complex reasoning steps or task execution processes.
For instance, Chain-of-InstructEditing \cite{zhang2023coie} harnesses this concept by generating sequential instructions to guide image editing tasks, illustrating how specific operations can refine the output for more precise editing. 
To avoid ambiguous user queries in table manipulation, \citet{zha2023tablegpt} introduce Chain-of-Command framework.
Inferring from user instructions, it enables LLMs to employ a series of pre-defined commands for more precise table execution. 
In the realm of e-commerce, \citet{li2023ecomgpt} implement a similar approach Chain-of-Task, which breaks down customer interactions into manageable atomic tasks with domain-specific e-commerce instructions, significantly simplifying complex user queries.

Recently, \citet{anugrah2024chain} propose Chain-of-Instructions. 
Different from previous methods using pre-defined or human-crafted instructions, this framework iteratively uses outputs of previous steps as instructions for the next step. 
An instruction dataset generated is then used for fine-tuning LLMs to handle complex instructions composed of multiple subtasks.
The results show that stepwise guidance can effectively improve the process and the outcomes of complex problem-solving tasks.

\paragraph{Histories}
Augmenting LLMs with historical data is essential for predictive modeling, which introduces another facet of Chain-of-Augmentation drawing contextual insights from the past. 
This approach is exemplified by Chain-of-Opinion \cite{do2023choire}, which analyzes historical user opinions to predict future reactions, offering valuable foresight into user sentiments.
In user-interface exploration, Chain-of-Action$^a$ \cite{zhan2023you} framework leverages past actions to guide future interactions, thereby optimizing user experience through predicted behaviors. 
\citet{ma2023large} take a similar approach in gaming environments like StarCraft II, where Chain-of-Summarization framework provides strategic gameplay recommendations based on a synthesis of past observations.

In addition to user modeling, the prediction of taxonomy structures also benefits from historical data, as seen in Chain-of-Layer \cite{zeng2024chain}, which builds upon previously identified categories. 
Temporal knowledge graphs receive a forward-looking treatment as well with methods like Chain-of-History$^a$ \cite{luo2024chain} and Chain-of-History$^b$ \cite{xia2024enhancing}, where historical graph structures inform LLM predictions about future nodal linkages or interactions.

\paragraph{Retrievals}
As the knowledge learned from pre-training data is limited and often outdated, LLMs frequently need to acquire external knowledge. 
Therefore, retrieval has become a crucial aspect of Chain-of-Augmentation. 
As one-step retrieval is often insufficient for complex tasks, methods have been developed to intersperse reasoning chains with explicit retrievals, thereby enhancing the quality of answers.
For example, ReAct \cite{yao2023react} synergizes reasoning and acting by adaptively retrieving external knowledge to augment the reasoning chains. 
While ReAct prompts LLMs to make decisions, Verify-and-Edit \cite{zhao-etal-2023-verify} decides when to retrieve based on less-than-majority agreement consistency and corrects erroneous reasoning chains to produce a more interpretable CoT.
Further refining this concept, \citet{li2023chain_k} develop Chain-of-Knowledge$^a$, which dynamically pulls relevant information from both unstructured and structured knowledge sources, e.g., Wikidata and tables. 
While ReAct and Verify-and-Edit keep all retrieved information in the chain, Chain-of-Knowledge$^a$ makes progressive corrections and only incorporates verified retrieval results to avoid propagating misleading information. 

\begin{table*}[!t]
    \centering
    \tabcolsep=0.13cm
    \fontsize{8pt}{10pt}\selectfont
    \renewcommand{\arraystretch}{1.0}
    \begin{tabular}{lcccc}
    \toprule \textbf{Chain-of-Retrievals} & \textbf{Self-Generated Query} & \textbf{Adaptive Retrieval} & \textbf{Retrieval Verification} & \textbf{Knowledge Sources} \\ \hline
    Self-Ask \cite{press-etal-2023-measuring} & \ding{51} & & & Textual Corpus \\
    ReAct \cite{yao2023react} & \ding{51} & \ding{51} &  & Textual Corpus \\
    Verify-and-Edit \cite{zhao-etal-2023-verify} & \ding{51} & \ding{51} & & Textual Corpus \\
    CoKnowledge$^a$ \cite{li2023chain_k} & & \ding{51} & \ding{51} & Textual \& Tabular Data \\
    IRCoT \cite{trivedi-etal-2023-interleaving} & \ding{51} & & & Textual Corpus \\
    CoQuery \cite{xu2023search} & \ding{51} & & \ding{51} & Textual Corpus \\
    CoAction$^b$ \cite{pan2024chain} & \ding{51} & & \ding{51} & Textual \& Market Data  \\
    RAT \cite{wang2024rat} & \ding{51} & & & Textual Corpus  \\
    ToG \cite{sun2024thinkongraph} & & \ding{51} & \ding{51} & Knowledge Graphs \\
    GraphCoT \cite{jin2024graph} & \ding{51} & \ding{51} & & Knowledge Graphs
    \\ \bottomrule
    \end{tabular}
    \renewcommand{\arraystretch}{1.0}
    \caption{A comparison of representative Chain-of-Retrievals methods from method and data source perspectives.}
    \vspace{-1em}
    \label{tab:CoRetrieval}
\end{table*}

Different from the adaptive retrieval in previous methods, another line of work explores how to compose informative queries during intermediate steps for knowledge retrieval. 
\citet{press-etal-2023-measuring} propose Self-Ask, prompting LLMs to ask follow-up questions themselves and answer these sub-questions with a Google Search API before generating the final response. 
Similarly, IRCoT \cite{trivedi-etal-2023-interleaving}, Chain-of-Question \cite{huang2024coq}, and RAT \cite{wang2024rat} augment each intermediate step with retrieved external knowledge iteratively refining the generations.
These methods directly insert retrievals into the reasoning chains, where LLMs can only reason about a local sub-question in each generation. 
Thus, when there is a misleading sub-question, the entire reasoning chain afterwards will be affected. 
To address this limitation and ensure the coherence of the global chain, \citet{xu2023search} develop the Chain-of-Query framework, which can interactively revisit previous retrievals and make necessary reasoning direction adjustments by verifying retrieved results.
Based on methodology designs and retrieval sources, we make further comparisons of previously discussed methods in Table \ref{tab:CoRetrieval}.


\paragraph{Tools}
Besides deploying a retriever to access external knowledge, recent methods have also explored utilizing other domain specific tools.
MultiToolCoT \cite{inaba-etal-2023-multitool} specifies available tools in the prompt and provides demonstration examples to enable tool using during CoT prompting.
To achieve more natural tool using, ChatCoT \cite{chen-etal-2023-chatcot} models CoT reasoning as multi-turn conversations, enabling LLMs to freely interact with tools through chatting.
To further improve the efficiency of interconnected tool calls, \citet{gao2024efficient} develop Chain-of-Abstraction, training LLMs to decode reasoning chains with abstract placeholders to be filled with knowledge from tools.
Such abstract chains enable LLMs to perform decoding and calling of external tools in parallel, thus reducing the inference delay of tool responses.

\paragraph{Others}
Other domain-specific augmentation methods have also been explored, e.g., Chain-of-Empathy \cite{lee2023chain} for empathetic response generation, Chain-of-Reference \cite{kuppa2023chain} for complex legal inquiries, and Chain-of-Dictionary \cite{lu2023chain} for machine translation.
These methods not only broaden the operational scope of LLMs but also underscore the potential of domain-specific CoX enhancements.

\subsection{Chain-of-Feedback}\label{sec:fdb}
Chain-of-Feedback represents another variant of CoX. 
Unlike augmentation which typically precedes generation, feedback is interlaced throughout the generation process to enhance and fine-tune responses. 
Based on the feedback source, we categorize them as external and self feedback.

\paragraph{External Feedback}
External feedback provides valuable perspectives overlooked by LLMs themselves. 
For instance, to generate 3D objects that LLMs may not have seen before, Chain-of-3DThought \cite{yamada2024l3go} utilizes external critiques to help iteratively hone an LLM’s understanding of 3D spaces for unconventional objects.
In addition to model feedback, human feedback is another important type of external feedback especially towards aligning LLMs with human preferences. 
Despite the success of RLHF \cite{ouyang2022training}, Chain-of-Hindsight \cite{liu2024chain} further transforms direct human preference data into natural language feedback that better aligns with how LLMs process textual information. 
Such feedback allows for more precise refinement to model's outputs, ensuring that responses are both accurate and contextually appropriate.

\paragraph{Self Feedback}
Though external feedback is critical, its costs and potential unavailability have led to a growing interest in self-refinement approaches \cite{lee2023rlaif}. 
Highlighted by \citet{madaan2023selfrefine}, Self-Refine first generates an initial response using an LLM and then uses the same LLM to provide feedback and refine its own response iteratively. 
Without any additional training, Self-Refine generates considerably better responses than direct generation.
Echoing this approach, \citet{dhuliawala2023chain} introduce Chain-of-Verification.
Instead of directly asking LLMs to provide feedback on their own responses, Chain-of-Verification asks LLMs to first plan a series of verification questions based on the initial responses and then answer these questions themselves.
After the self-assessment, LLMs are then asked to generate their final verified answers.
Chain-of-NLI \cite{lei2023chain} adopts similar framework but formulates a series of natural language inference problems to be answered. 
Similar concept has also been applied in other tasks. 
Chain-of-Density \cite{adams-etal-2023-sparse} allows LLMs to iteratively refine generated summaries by incorporating self-detected missing information.
Chain-of-SelfRevisions \cite{le2024codechain} improves modular code generation by reusing previously generated code modules. 

\begin{figure}[!t]
    \centering
    \includegraphics[width=0.83\linewidth]{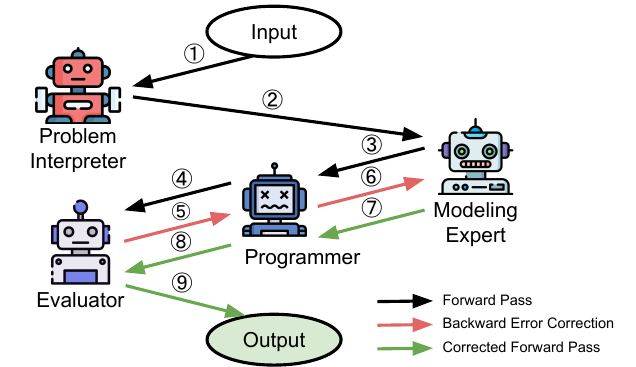}
    \caption{A simplified illustrative workflow of Chain-of-Experts \cite{xiao2024chainofexperts}.}
    \vspace{-0.5em}
    \label{fig:coe}
\end{figure}

\subsection{Chain-of-Models}\label{sec:models}
Previous CoX methods have mostly been designed for a single LLM. 
Recognizing that different LLMs may have different specialties \cite{xiao2024chainofexperts, xia2024llm}, another line of work proposes constructing a chain of models to leverage distinct strengths of each model.
Chain-of-Experts \cite{xiao2024chainofexperts} exemplifies this collaborative strategy. 
As illustrated in Figure \ref{fig:coe}, it involves a consortium of expert LLMs that work in sequence, each contributing its specialized knowledge to build upon the results developed by its predecessors. 
This method is particularly effective in addressing intricate problems in operations research, where the complexity often exceeds the processing capabilities of a single LLM.
Similarly, \citet{qiu2024chain} deploy a chain of specialized LoRA \cite{hu2022lora} networks, each fine-tuned to effectively handle different domains of a broader problem. 
This approach ensures that specific tasks benefit from the most relevant expertise, enhancing overall efficiency and outcome accuracy.
In parallel, ChatEval \cite{chan2024chateval} and Chain-of-Discussion \cite{tao2024chain} employ multiple LLMs engaging in a dialogue, critiquing and refining each other’s contributions before reaching a consensus in the final response. 
This process ensures that the synthesized output is not only comprehensive but also critically evaluated from multiple perspectives.

\section{Chain-of-X Tasks}\label{sec:tasks}
CoX can be of various forms, enabling their applications in diverse areas. 
This section surveys CoX methods categorized by tasks as shown in Figure \ref{fig:survey}.

\subsection{Multi-Modal Interaction}\label{sec:mm_inter}
Real-world problem-solving tasks often involve modalities other than text. Thus, the success of CoT has drawn attention to designing frontier CoX methods for challenges in multi-modality.

\paragraph{Text-Image}
To handle rich features from both texts and images, the knowledge composition ability of CoX methods has been crucial in capturing key information. 
While MultimodalCoT \cite{zhang2023multimodal} incorporates image information into textual rationale generation, it lacks interaction between modalities.
To address this, methods have explored using intermediate reasoning steps to infer and extract key visual information before generating final responses. 
For example, DDCoT \cite{zheng2023ddcot} utilizes structured logical chains to explicitly guide the understanding of relevant image regions.
Chain-of-Look \cite{xi2023chain} constructs a visual semantic reasoning chain based on textual cues for visual entity recognition.
Chain-of-Manipulation \cite{qi2024cogcom} and Chain-of-Spot \cite{liu2024chain-a} adopt a step-wise refinement process for identifying critical image details. 

\paragraph{Text-Table}
Unlike texts, structured tabular data has been a challenging source for LLMs to reason with or manipulate \cite{fang2024large}. 
To this end, CoX methods show advantages in decomposing complex table operations into a sequence of manageable subtasks.
\citet{zha2023tablegpt} utilize a sequence of pre-defined commands to execute table operations step by step until the queried information is found. 
Taking a step further, Chain-of-Table \cite{wang2023chain} directly leverages tabular data as a part of the reasoning chain. 
Here, tables are not just data sources but act as evolving entities within the reasoning process, dynamically being updated in response to the LLM’s queries and tasks. 
This iterative process allows the model to engage with the table more naturally and effectively, leading to a more nuanced understanding and manipulation of table information.

\paragraph{Text-Code}
The nature of sequential execution of code generation makes it another task benefiting from CoX methods \cite{zan-etal-2023-large}. 
Chain-of-Repair \cite{wang2023intervenor} draws inspiration from traditional debugging processes.
It employs a teacher model to interpret compiler feedback and compose a chain of code repairing steps, teaching a student model to generate debugged code.
Chain-of-SelfRevisions \cite{le2024codechain} explores modular code generation. 
This method iteratively extracts and clusters sub-modules from previous generations and adds them to new reasoning chains, naturally encouraging code reuse and efficiency.

\paragraph{Text-Speech}
The field of speech generation has also seen innovative applications of CoX methods.
Chain-of-Information \cite{zhang2024speechgpt} enhances speech synthesis by separating and then reassembling semantic and perceptual components, which allows for more nuanced and accurate speech output.
Chain-of-Modality \cite{zhang-etal-2023-speechgpt} merges textual and vocal instructions to guide speech generation. 
This method not only enhances the quality of speech generation but also enables LLMs to handle conversational nuances, effectively bridging the gap between textual and speech data. 

\subsection{Factuality \& Safety}\label{sec:fact}
Ensuring factuality and safety in LLM outputs has been critical for practical applications \cite{wang2023aligning}.
Recent studies have also explored CoX for both hallucination reduction and alignment. 

\paragraph{Hallucination Reduction}
LLMs have shown a propensity for generating hallucinations \cite{agrawal2023can, xia-etal-2024-hallucination}.
Two main sources of hallucinations include: i) LLMs being overconfident in their incorrect understanding of the problem and overlooking details, and ii) LLMs having limited knowledge about certain tasks and generating uncertain answers \cite{zhang2023siren}.
For the first type, step-wise verification and iterative refinement have been applied to guide LLMs to reassess their initial response and focus on details, exemplified by Self-Refine \cite{madaan2023selfrefine}, Chain-of-NLI \cite{lei2023chain}, and Chain-of-Verification \cite{dhuliawala2023chain}.
For the second type, it is essential to augment LLMs with additional knowledge grounding their responses. 
Several CoX methods, e.g., Chain-of-Note \cite{yu2023chain}, Chain-of-Knowledge$^a$ \cite{li2023chain_k}, and Chain-of-Action$^b$ \cite{pan2024chain} retrieve domain-specific knowledge at each step, effectively reducing the occurrence of unfactual generations.

\paragraph{Alignment}
Aligning LLMs with human preferences is critical to ensure that LLMs generate helpful and harmless responses.
Despite the wide adoption of RLHF for LLM alignment \cite{ouyang2022training, xia-etal-2024-aligning}, there are still challenges including the high cost of human annotation and imperfect reward functions.
To help LLMs learn from any feedback form, \citet{liu2024chain} proposes transforming preference data into a sequence of natural language sentences for supervised fine-tuning.
Leveraging the language comprehension capabilities of LLMs, Chain-of-Hindsight achieves better alignment performance compared to RLHF.
Meanwhile, Chain-of-Utterance prompting \cite{bhardwaj2023red} has been proposed for LLM red-teaming.
It adopts a sequential structure to establish a jailbreaking conversation between a harmful LLM and a helpful but unsafe LLM, exposing safety issues of LLMs to be aligned.

\subsection{Multi-Step Reasoning}\label{sec:ms_reas}
Multi-step reasoning typically demands a robust understanding of context and logic \cite{wei2022chain}. 
These tasks require breaking down complex problems into a series of smaller, interconnected steps, building upon each step to reach a logical conclusion. 
The sequential nature of CoX makes it ideally suited for these tasks, including
rule-based reasoning \cite{servantez2024chain}, database reasoning \cite{hu2023chatdb}, legal reasoning \cite{kuppa2023chain}, user behavior reasoning \cite{do2023choire, han2024chain}, graph reasoning \cite{zeng2024chain, luo2024chain, xia2024enhancing}, as well as reasoning for summarization \cite{adams-etal-2023-sparse, bao2024chain} and machine translation \cite{lu2023chain}.
These varied applications demonstrate CoX's advantages in enhancing LLMs' ability to process information more effectively.

\subsection{Instruction Following}\label{sec:instr_fl}
Instruction following has been a celebrated ability of LLMs \cite{zhang2023instruction}.
The evolution of CoX methods has also led to various approaches for enhancing this feature.
A notable line of works, such as Chain-of-Task \cite{li2023ecomgpt}, LogiCoT \cite{liu-etal-2023-logicot} and Chain-of-Imagination \cite{zhou2024minedreamer}, construct sequences of instructions for prompting or instruction tuning to handle complex tasks that require explicit step-wise guidance.
While training a single LLM to follow different instructions can be costly, Chain-of-LoRA \cite{qiu2024chain} adopts a series of LoRA networks to specialize in instruction handling.
After identifying an instruction type, Chain-of-LoRA applies task-specific LoRA networks to the base LLM to accomplish the respective tasks. 

\subsection{LLMs as Agents}\label{sec:agent}
The planning abilities have made LLMs strong agents across a wide range of tasks \cite{xi2023rise}.
CoX methods have been explored to further boost the planning abilities of LLM-based agents.
In this vein, Chain-of-Action$^a$ \cite{zhan2023you} and Chain-of-ActionThought \cite{zhang2024android} utilize a series of planned actions to guide the decision-making of agents, ensuring each step is informed by the previous actions. 
As discussed previously, LLM-based agents can also be augmented with historical data, e.g., Chain-of-Summarization \cite{ma2023large}, and external model feedback, e.g., Chain-of-3DThought \cite{yamada2024l3go}.
LLMs also serve as planners in human-scene interaction tasks with Chain-of-Contacts \cite{xiao2024unified}, and in tool using with Chain-of-Abstraction \cite{gao2024efficient}.
Chain-of-Models also naturally involves multi-agent settings such as Chain-of-Discussion \cite{tao2024chain}. 
These methods highlight the integration of CoX in enhancing the multi-dimensional abilities of LLMs as autonomous and collaborative agents.

\subsection{Evaluation Tools}\label{sec:eval}
Evaluating LLMs has become increasingly challenging as they grow more sophisticated \cite{chang2023survey}, making CoX methods a valuable asset for evaluation purposes.
Chain-of-Utterances prompting \cite{bhardwaj2023red} exposes safety issues where LLMs interact with potentially harmful models. 
BadChain \cite{xiang2024badchain} also reveals vulnerabilities of LLMs outputting unintended malicious content under backdoor attack when employing CoT prompting.
Chain-of-Feedback \cite{ahn2024chain} conducts another evaluation, demonstrating that by repeatedly providing LLMs with non-informative prompts like ``\textit{make another attempt}'', the quality of responses gradually decreases. 
These methods underscore the importance of nuanced evaluations of LLMs.
\section{Future Directions}\label{sec:discussion}

While LLMs have demonstrated remarkable abilities in step-by-step problem-solving for various tasks, several challenges remain to be addressed.


\paragraph{Causal Analysis on Intermediates}
Existing works generally focus on improving task-specific generative results. However, understanding and explaining the underlying mechanisms of LLM reasoning is also essential in realistic scenarios.
For example, \citet{wang2023selfconsistency} show that LLMs may skip rational steps when generating final results. \citet{wang-etal-2023-towards} observe a performance gain from CoT even with invalid rationales. These observations indicate the value of a causal analysis on how intermediate steps truly affect the final results.

\paragraph{Reducing Inference Cost}
A chain leading to the final node of generation often requires multiple sequential inference steps, which are computationally heavy and time-consuming, especially with LLMs.
Future research may explore reducing the length of CoX chains while maintaining the quality of generation.
It would also be worth studying whether the intermediate nodes of CoX could be executed in parallel or jointly within a single inference step.


\paragraph{Knowledge Distillation}
The knowledge elicited by the intermediate nodes of CoX contains fine-grained task instructions, which can benefit the training of smaller student models when using a teacher LLM for knowledge distillation.
\citet{li-etal-2023-symbolic} and \citet{hsieh-etal-2023-distilling} have shown that the student model can effectively learn from the rationales of CoT generated by an LLM.
Nonetheless, it remains an open question whether the intermediate nodes from broader CoX methods are equally informative in inspiring student learning.

\paragraph{End-to-End Fine-tuning}
One drawback of CoX is that it does not follow an end-to-end paradigm; i.e., generation errors may accumulate along the chain when self-correction \cite{le2024codechain, dhuliawala2023chain} is not enforced. 
Future research can explore fine-tuning LLMs with CoX prompting and penalizing errors from the final output. By reducing the generation errors end-to-end, we expect this will improve the quality of both the intermediate and final nodes in CoX.

\section{Conclusion}
This survey explored Chain-of-X methods, building upon the concept of Chain-of-Thought. 
By categorizing them based on nodes and tasks, we provide a comprehensive overview that highlights the potential of CoX in enhancing LLM capabilities and opens new avenues for future research. 
Through this survey, we aim to inspire further exploration in a deeper understanding and more creative use of CoX paradigms for LLMs.

\newpage
\section*{Limitations}
This survey presents a comprehensive overview of Chain-of-X methods. 
We have made our best efforts to collect studies leveraging the CoX concept, regardless of whether they are explicitly named as such. 
However, with the rapidly growing number of works in the field, there is still a chance that some have been missed. 
We welcome suggestions from the research community and will continue our efforts to survey and update the collected works.




\bibliography{anthology,custom}

\appendix

\section{Taxonomies of Nodes and Tasks}\label{sec:app}
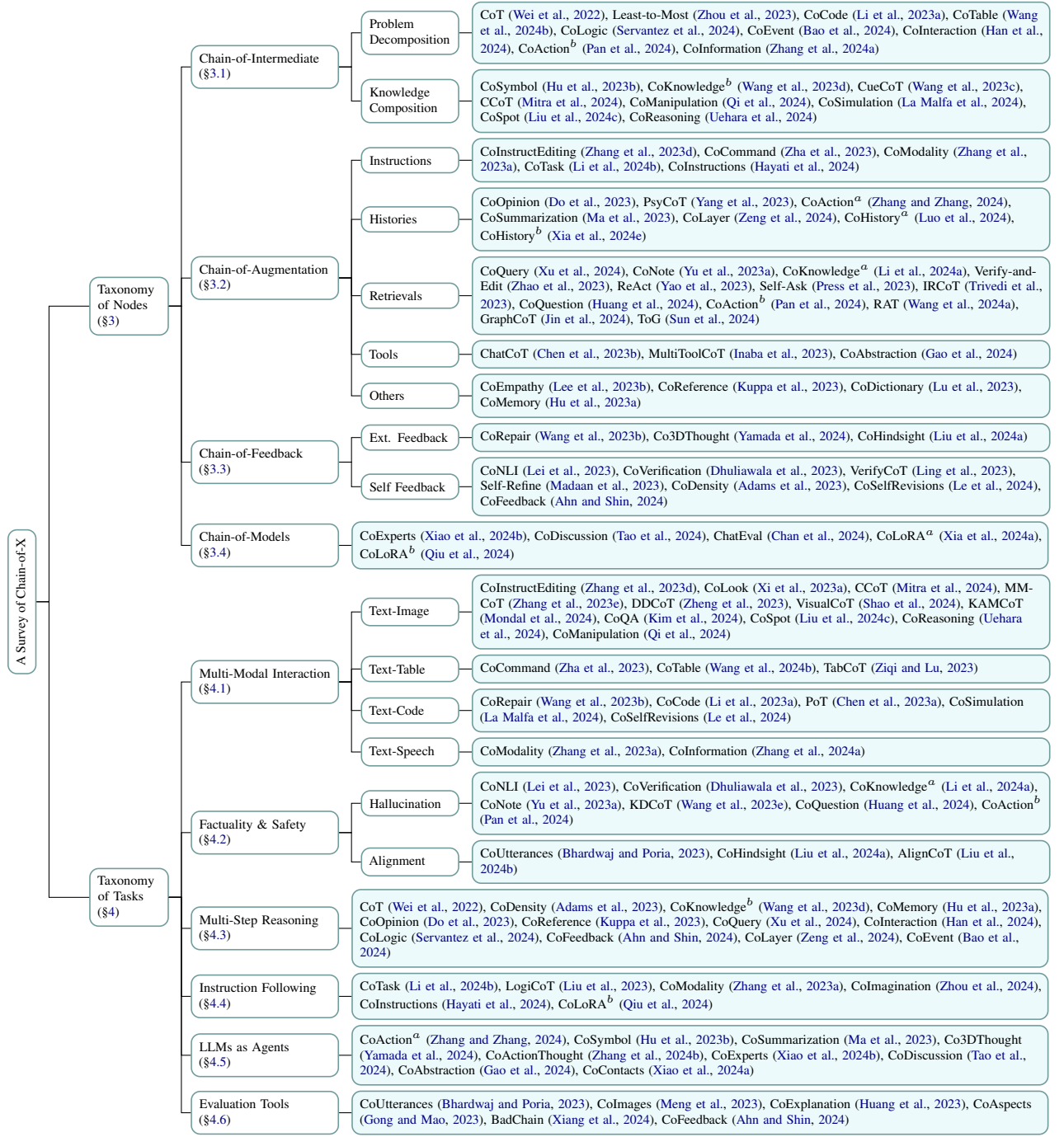
\begin{figure*}[t!]
\centering
\begin{forest}
for tree={   
font=\tiny, 
draw=myblue, semithick, rounded corners,
       minimum height = 1.5ex,
        minimum width = 2em,
    anchor = west,
     grow = east,
forked edge,        
    s sep = 0.8mm,    
    l sep = 3.5mm,    
 fork sep = 2mm,    
           }
[A Survey of Chain-of-X, rotate=90, anchor=center
    [Taxonomy of Tasks (\S \ref{sec:tasks}), fit=band, text width=1cm
        [Evaluation Tools \\ (\S \ref{sec:eval}), text width=2.1cm, l sep = 2mm
            [{CoUtterances \cite{bhardwaj2023red}, CoImages \cite{meng2023chain}, CoExplanation \cite{huang2023chain}, CoAspects \cite{gong2023coascore}, BadChain \cite{xiang2024badchain}, CoFeedback \cite{ahn2024chain}}, text width=10.9cm, fill=lightblue]
        ]
        [LLMs as Agents \\ (\S \ref{sec:agent}), text width=2.1cm, l sep = 2mm
            [{CoAction$^a$ \cite{zhan2023you}, CoSymbol \cite{hu2023chain}, CoSummarization \cite{ma2023large}, Co3DThought \cite{yamada2024l3go}, CoActionThought \cite{zhang2024android}, CoExperts \cite{xiao2024chainofexperts}, CoDiscussion \cite{tao2024chain}, CoAbstraction \cite{gao2024efficient}, CoContacts \cite{xiao2024unified}}, text width=10.9cm, fill=lightblue]
        ]
        [Instruction Following (\S \ref{sec:instr_fl}), text width=2.1cm, l sep = 2mm
            [{CoTask \cite{li2023ecomgpt}, LogiCoT \cite{liu-etal-2023-logicot}, CoModality \cite{zhang-etal-2023-speechgpt}, CoImagination \cite{zhou2024minedreamer}, CoInstructions \cite{anugrah2024chain}, CoLoRA$^b$ \cite{qiu2024chain}}, text width=10.9cm, fill=lightblue]
        ]
        [Multi-Step Reasoning (\S \ref{sec:ms_reas}), text width=2.1cm, l sep = 2mm
            [{CoT \cite{wei2022chain}, CoDensity \cite{adams-etal-2023-sparse}, CoKnowledge$^b$ \cite{wang2023boosting}, CoMemory \cite{hu2023chatdb}, CoOpinion \cite{do2023choire}, CoReference \cite{kuppa2023chain}, CoQuery \cite{xu2023search}, CoInteraction \cite{han2024chain}, CoLogic \cite{servantez2024chain}, CoFeedback \cite{ahn2024chain}, CoLayer \cite{zeng2024chain}, CoEvent \cite{bao2024chain}}, text width=10.9cm, fill=lightblue]
        ]
        [Factuality \& Safety (\S \ref{sec:fact}), text width=2.1cm
            [Alignment, text width=1.3cm, l sep = 2mm
                [{CoUtterances \cite{bhardwaj2023red}, CoHindsight \cite{liu2024chain}, AlignCoT \cite{liu2024mixture}}, text width=9.0cm, fill=lightblue]
            ]
            [Hallucination, text width=1.3cm, l sep = 2mm
                [{CoNLI \cite{lei2023chain}, CoVerification \cite{dhuliawala2023chain}, CoKnowledge$^a$ \cite{li2023chain_k}, CoNote \cite{yu2023chain}, KDCoT \cite{wang2023knowledge}, CoQuestion \cite{huang2024coq}, CoAction$^b$ \cite{pan2024chain}}, text width=9.0cm, fill=lightblue]
            ]
        ]
        [Multi-Modal Interaction (\S \ref{sec:mm_inter}), text width=2.1cm
            [Text-Speech, text width=1.3cm, l sep = 2mm
                [{CoModality \cite{zhang-etal-2023-speechgpt}, CoInformation \cite{zhang2024speechgpt}}, text width=9.0cm, fill=lightblue]
            ]
            [Text-Code, text width=1.3cm, l sep = 2mm
                [{CoRepair \cite{wang2023intervenor}, CoCode \cite{li2023chain}, PoT \cite{chen2023program}, CoSimulation \cite{la2024code}, CoSelfRevisions \cite{le2024codechain}}, text width=9.0cm, fill=lightblue]
            ]
            [Text-Table, text width=1.3cm, l sep = 2mm
                [{CoCommand \cite{zha2023tablegpt}, CoTable \cite{wang2023chain}, TabCoT \cite{ziqi-lu-2023-tab}}, text width=9.0cm, fill=lightblue]
            ]
            [Text-Image, text width=1.3cm, l sep = 2mm
                [{CoInstructEditing \cite{zhang2023coie}, CoLook \cite{xi2023chain}, CCoT \cite{mitra2023compositional}, MMCoT \cite{zhang2023multimodal}, DDCoT \cite{zheng2023ddcot}, VisualCoT \cite{shao2024visual}, KAMCoT \cite{Mondal2024KAMCoTKA}, CoQA \cite{kim2024generalizing}, CoSpot \cite{liu2024chain-a}, CoReasoning \cite{uehara2024advancing}, CoManipulation \cite{qi2024cogcom}}, text width=9.0cm, fill=lightblue]
            ]
        ]
    ]
    [Taxonomy of Nodes (\S \ref{sec:nodes}), fit=band, text width=1cm
        [Chain-of-Models \\ (\S \ref{sec:models}), text width=2.1cm, l sep = 2mm
            [{CoExperts \cite{xiao2024chainofexperts}, CoDiscussion \cite{tao2024chain}, 
            ChatEval \cite{chan2024chateval}, CoLoRA$^a$ \cite{xia2024chain}, CoLoRA$^b$ \cite{qiu2024chain}}, text width=10.9cm, fill=lightblue]
        ]
        [Chain-of-Feedback (\S \ref{sec:fdb}), text width=2.1cm
            [Self Feedback, text width=1.3cm, l sep = 2mm
                [{CoNLI \cite{lei2023chain}, CoVerification \cite{dhuliawala2023chain}, VerifyCoT \cite{ling2023deductive}, Self-Refine \cite{madaan2023selfrefine}, CoDensity \cite{adams-etal-2023-sparse}, CoSelfRevisions \cite{le2024codechain}, CoFeedback \cite{ahn2024chain}}, text width=9.0cm, fill=lightblue]
            ]
            [Ext. Feedback, text width=1.3cm, l sep = 2mm
                [{CoRepair \cite{wang2023intervenor}, Co3DThought \cite{yamada2024l3go}, CoHindsight \cite{liu2024chain}}, text width=9.0cm, fill=lightblue]
            ]
        ]
        [Chain-of-Augmentation (\S \ref{sec:augmt}), text width=2.1cm
            [Others, text width=1.3cm, l sep = 2mm
                [{CoEmpathy \cite{lee2023chain}, CoReference \cite{kuppa2023chain}, CoDictionary \cite{lu2023chain}, CoMemory \cite{hu2023chatdb}}, text width=9.0cm, fill=lightblue]
            ]
            [Tools, text width=1.3cm, l sep = 2mm
                [{ChatCoT \cite{chen-etal-2023-chatcot},  MultiToolCoT \cite{inaba-etal-2023-multitool}, CoAbstraction \cite{gao2024efficient}}, text width=9.0cm, fill=lightblue]
            ]
            [Retrievals, text width=1.3cm, l sep = 2mm
                [{CoQuery \cite{xu2023search}, CoNote \cite{yu2023chain}, CoKnowledge$^a$ \cite{li2023chain_k}, Verify-and-Edit \cite{zhao-etal-2023-verify}, ReAct \cite{yao2023react}, Self-Ask \cite{press-etal-2023-measuring}, IRCoT \cite{trivedi-etal-2023-interleaving}, CoQuestion \cite{huang2024coq}, CoAction$^b$ \cite{pan2024chain}, RAT \cite{wang2024rat}, GraphCoT \cite{jin2024graph}, ToG \cite{sun2024thinkongraph}}, text width=9.0cm, fill=lightblue]
            ]
            [Histories, text width=1.3cm, l sep = 2mm
                [{CoOpinion \cite{do2023choire}, PsyCoT \cite{yang-etal-2023-psycot}, CoAction$^a$ \cite{zhan2023you}, CoSummarization \cite{ma2023large}, CoLayer \cite{zeng2024chain}, CoHistory$^a$ \cite{luo2024chain}, CoHistory$^b$ \cite{xia2024enhancing}}, text width=9.0cm, fill=lightblue]
            ]
            [Instructions, text width=1.3cm, l sep = 2mm
                [{CoInstructEditing \cite{zhang2023coie}, CoCommand \cite{zha2023tablegpt}, CoModality \cite{zhang-etal-2023-speechgpt}, CoTask \cite{li2023ecomgpt}, CoInstructions \cite{anugrah2024chain}}, text width=9.0cm, fill=lightblue]
            ]
        ]
        [Chain-of-Intermediate (\S \ref{sec:inter}), text width=2.1cm
            [Knowledge \\ Composition, text width=1.3cm, l sep = 2mm
                [{CoSymbol \cite{hu2023chain}, CoKnowledge$^b$ \cite{wang2023boosting}, CueCoT \cite{wang-etal-2023-cue}, CCoT \cite{mitra2023compositional}, CoManipulation \cite{qi2024cogcom}, CoSimulation \cite{la2024code}, CoSpot \cite{liu2024chain-a}, CoReasoning \cite{uehara2024advancing}}, text width=9.0cm, fill=lightblue]
            ]
            [Problem \\ Decomposition, text width=1.3cm, l sep = 2mm
                [{CoT \cite{wei2022chain}, Least-to-Most \cite{zhou2023leasttomost}, CoCode \cite{li2023chain}, CoTable \cite{wang2023chain}, CoLogic \cite{servantez2024chain}, CoEvent \cite{bao2024chain}, CoInteraction \cite{han2024chain}, CoAction$^b$ \cite{pan2024chain}, CoInformation \cite{zhang2024speechgpt}}, text width=9.0cm, fill=lightblue]
            ]
        ]
    ]
]
\end{forest}
\caption{A Survey of Chain-of-X by Taxonomies of Nodes and Tasks.}
\label{fig:app}
\end{figure*}

We present in Figure \ref{fig:app} the complete version of Figure \ref{fig:survey} on Chain-of-X taxonomies categorized by nodes and tasks.

\end{document}